\documentclass{article}

\usepackage[thai, main=english]{babel}
\usepackage{ucs}
\usepackage[utf8x]{inputenc}
\usepackage[T1]{fontenc}
\usepackage{verbatim}
\usepackage{amsmath}
\usepackage{graphicx}
\usepackage{float}
\usepackage{tabularray}
\usepackage{multirow}
\usepackage{array}
\usepackage{longtable}
\usepackage[colorlinks=true, allcolors=blue]{hyperref}
\usepackage[letterpaper,top=2cm,bottom=2cm,left=3cm,right=3cm,marginparwidth=1.75cm]{geometry}
\SetUnicodeOption{default}

\newcolumntype{L}[1]{>{\raggedright\arraybackslash}p{#1}}

\title{Thai Financial Domain Adaptation of THaLLE --\\Technical Report}
\author{NLP-Voice Research Lab, KBTG Labs,\\
\textit{KASIKORN Business---Technology Group}\\}

\begin{document}
\maketitle

\begin{abstract}
Large Language Models (LLMs) excel in general tasks but struggle with domain-specific challenges, such as specialized terminology and localized regulations. Existing financial LLMs, like FinGPT and BloombergGPT, lack support for the Thai financial domain.
We developed a Thai Financial LLM using the Investment Consultant (IC) exam dataset from the Stock Exchange of Thailand. To address dataset limitations, we applied data augmentation, ReLoRA for efficient training, Continued Pretraining (CPT) for domain knowledge, and Rank-Stabilized LoRA (rsLoRA) for fine-tuning. Supervised Fine-Tuning (SFT) simulated exam scenarios, while Direct Preference Optimization (DPO) refined the model using feedback.
The model achieved scores of 72\%, 72\%, and 84\% on IC exam levels P1, P2, and P3, respectively, demonstrating its effectiveness in Thai financial advisory tasks and its potential for specialized applications.
\end{abstract}

\section{Introduction}
\label{sec:introduction}

In recent years, Large Language Models (LLMs) have demonstrated remarkable improvements across a variety of tasks, particularly in conversational system. These models are designed to understand and learn general knowledge from their training data, which often includes diverse and extensive text sources. Notable among them is Llama 3.1 \cite{dubey2024llama3herdmodels}, which has achieved significant performance in conversational tasks, including providing accurate answers even without additional context. Similarly, numerous other base models, trained on varied datasets and fine-tuned with different methodologies, have been introduced, showcasing impressive capabilities to follow instructions \cite{qwen2.5,geminiteam2024gemini15unlockingmultimodal,openai2024gpt4technicalreport}.

Despite these achievements, LLMs face notable challenges when dealing with domain-specific knowledge. They often struggle with specialized terminology, abbreviations, and nuanced concepts unique to particular fields. Furthermore, their grounding and reasoning capabilities are typically built on general context and are not customized for specific scenarios. Financial use cases demands specialized knowledge and reasoning skills to address domain-specific queries. This includes understanding technical terms and abbreviations related to financial products (e.g., Retirement Mutual Fund (RMF)). Moreover, managing complex calculations and addressing localized contexts, such as country-specific regulatory frameworks, poses a considerable challenge.

Recent developments in financial language models have introduced significant advancements. Among the most notable are FinGPT, an open-source financial LLM designed for enhanced data accessibility and lightweight adaptation \cite{yang2023fingptopensourcefinanciallarge}, and BloombergGPT, a proprietary model trained on vast amounts of financial data to deliver exceptional performance in finance-specific tasks \cite{wu2023bloomberggptlargelanguagemodel}. However, these advances do not adequately transfer to financial applications in the Thai context. The market currently lacks a model that combines financial expertise, Thai language proficiency, and domain-specific financial knowledge tailored to Thailand's unique needs.

Building upon this gap, our work focuses on developing a language model specifically tailored for the Thai financial domain. To achieve this, we utilized the Investment Consultant (IC) exam issued by the Stock Exchange of Thailand (SET) as a foundational dataset. Given the dataset's limited size, data augmentation techniques were employed to expand its scope and diversity. To further enhance the model's learning capabilities, we adopted ReLoRA, a method that applies low-rank updates to train high-rank networks, improving both training efficiency and performance \cite{lialin2023relorahighranktraininglowrank}. The training process also incorporated CPT to establish robust foundational knowledge \cite{ke2023continualpretraininglanguagemodels}, while Rank-Stabilized LoRA (rsLoRA) was used for efficient fine-tuning \cite{kalajdzievski2023rankstabilizationscalingfactor}.

Moreover, to emulate realistic examination and learning scenarios, we implemented two fine-tuning objectives: SFT simulating exam conditions, and DPO refining the model by leveraging feedback from incorrect responses \cite{NEURIPS2023_a85b405e}. These methodologies collectively ensure the model's alignment with the unique requirements of the Thai financial domain while maintaining high training efficiency and adaptability.

As a result, our contributions are summarized as follows:

\begin{itemize}
    \item \textbf{Development of a Thai Financial LLM}: Created a language model specifically tailored for the Thai financial domain using the IC exam dataset provided by SET.
    \item \textbf{Data Augmentation for Limited Datasets}: Applied data augmentation techniques to address the constraints of a limited dataset, expanding the training corpus for better model performance.
    \item \textbf{Effective Training and Fine-Tuning Framework}: Employed ReLoRA for efficient and scalable training by applying low-rank updates to high-rank networks \cite{lialin2023relorahighranktraininglowrank}, utilized CPT to build foundational domain knowledge \cite{ke2023continualpretraininglanguagemodels}, and integrated rsLoRA for computationally efficient fine-tuning \cite{kalajdzievski2023rankstabilizationscalingfactor}.
    \item \textbf{Simulation of Examination and Feedback-Driven Refinement}: Applied SFT for exam scenario simulation and DPO to refine the model through feedback on incorrect responses \cite{NEURIPS2023_a85b405e}.
\end{itemize}

By integrating these advanced techniques and methodologies, we successfully developed an LLM specifically designed to navigate the complexities of financial advisory. The model demonstrates its effectiveness by passing all levels of the IC exam, achieving scores of 72\%, 72\%, and 84\% for P1, P2, and P3, respectively.

\section{Background}
\label{sec:background}

\subsection{Financial Domain LLMs}
 
LLMs provide robust foundations for financial applications through their pre-trained capabilities in diverse linguistic tasks \cite{dubey2024llama3herdmodels,yang2024qwen2technicalreport,damonlp2024seallm3,falcon40b,glm2024chatglmfamilylargelanguage,workshop2023bloom176bparameteropenaccessmultilingual,MosaicML2023Introducing}.
Each model brings unique strengths, including multilingual support, computational efficiency, and optimized architectures. However, their general-purpose design often necessitates domain-specific adaptation to address the specialized requirements of financial datasets and terminology.
 
Tailored financial LLMs have been developed to meet these challenges. FinBERT \cite{liu2021finbert} specializes in sentiment analysis within financial texts, leveraging a rich domain corpus to achieve high accuracy. FLUE and its derivative FLANG-BERT \cite{shah-etal-2022-flang} serve as benchmarks for financial language understanding, outperforming general-purpose models. BloombergGPT \cite{wu2023bloomberggptlargelanguagemodel} integrates proprietary financial datasets, offering comprehensive coverage of financial tasks, while FinGPT \cite{yang2023fingptopensourcefinanciallarge} advances accessibility through open-source low-rank adaptation techniques. PIXIU \cite{xie2023pixiu} contributes as a benchmark for evaluating financial LLMs, particularly those based on Llama models.

However, existing financial LLMs frequently fall short when it comes to country-specific knowledge, such as Thailand's unique regulations, financial systems, and terminology, which hinders their effectiveness. Developing tailored LLMs that incorporate this specialized knowledge is crucial for delivering accurate understanding and contextually appropriate responses.

\subsection{The Investment Consultant License Exam}
The IC license exam \cite{SET2024ICCourse}, issued by SET, is a mandatory certification for individuals seeking to professionally provide investment advice across a range of financial products. The exam is structured into three levels: P1, P2, and P3. P1 serves as the foundational level, while P2 and P3 progressively expand on the knowledge and skills introduced in P1. The details of each level are outlined as follows.

\subsubsection{Plain Product (P1)}
The Plain Product exam evaluates candidates on three core areas: Fundamental Knowledge, which includes investment environments, financial markets, risk and return, diversification, and various types of analyses; Related Rules and Regulations and Suitable Investment Consulting, focusing on professional conduct, regulatory compliance, and tailored investment advice; and Product Knowledge, covering equities, fixed income, mutual funds, and their associated valuation, risks, and trading mechanisms. Together, these modules ensure a comprehensive understanding of the investment landscape and the skills required for effective consulting.
\begin{itemize}
    \item \textbf{Format}: 100 multiple-choice questions with four options 
    \item \textbf{Duration}: 2 hours and 30 minutes 
    \item \textbf{Passing Criteria}: 70\% of the total score and 70\% in Related Rules and Regulations and Suitable Investment Consulting module
\end{itemize}

\subsubsection{Complex Product 1 (P2)}
The Complex Products: Bond and Mutual Fund exam assesses candidates on their understanding of high-risk and complex financial products. It includes knowledge of Complex Bonds, such as structured debt securities, non-investment-grade bonds, and unrated securities, focusing on their characteristics, yield calculations, and associated risks. Additionally, it covers Complex Mutual Funds, emphasizing the ability to differentiate between fund types, calculate returns, and evaluate risks. The exam also highlights Investment Consulting for Complex Bonds and Mutual Funds, ensuring candidates can provide informed advice through proper consulting practices, leveraging investment channels, and utilizing reliable information sources.
\begin{itemize}
    \item \textbf{Format}: 25 multiple-choice questions with four options 
    \item \textbf{Duration}: 40 minutes
    \item \textbf{Passing Criteria}: 70\% of the total score
\end{itemize}

\subsubsection{Complex Product 2 (P3)}
The Complex Products: Derivatives exam evaluates candidates on their understanding of derivatives through key areas, including the fundamentals of derivatives and their types, underlying assets, and market participants. It covers knowledge of futures contracts, focusing on pricing, valuation, and strategies for hedging, speculation, and arbitrage, as well as options contracts, emphasizing their definitions, pricing factors, and strategic uses. The exam also addresses the trading mechanisms of the Thailand Futures Exchange (TFEX), including trading procedures, clearing, and settlement processes, alongside detailed knowledge of the contract specifications for derivative products traded on TFEX.
\begin{itemize}
    \item \textbf{Format}: 50 multiple-choice questions with four options 
    \item \textbf{Duration}: 1 hour and 20 minutes 
    \item \textbf{Passing Criteria}: 70\% of the total score
\end{itemize}

\subsection{ReLoRA: High-Rank Training Through Low-Rank Updates}

ReLoRA \cite{lialin2023relorahighranktraininglowrank} is a parameter-efficient training technique designed to train large neural networks by employing low-rank updates to achieve high-rank performance. This method builds upon the Low-Rank Adaptation (LoRA) approach, which introduces low-rank matrices to fine-tune pre-trained models with reduced computational overhead. While LoRA is effective for fine-tuning, ReLoRA extends its applicability to training models from scratch.

The core idea of ReLoRA is to perform a series of low-rank updates that, when aggregated, approximate a high-rank update. This is based on the mathematical property that the rank of the sum of two matrices is less than or equal to the sum of their ranks. By iteratively applying low-rank updates and merging them into the original model parameters, ReLoRA incrementally increases the effective rank of the model's weight matrices. This process involves several key components:

Initial Full-Rank Training: A brief phase of full-rank training to ``warm start'' the model, establishing a solid foundation for subsequent low-rank updates.
LoRA Training with Restarts: Implementing LoRA-based low-rank updates, followed by periodic restarts where the low-rank matrices are merged into the main model parameters.
Jagged Learning Rate Schedule: Utilizing a learning rate schedule that resets to zero after each restart, followed by a quick warm-up, to stabilize training dynamics.
Partial Optimizer Resets: Resetting parts of the optimizer state during restarts to prevent the optimizer's momentum from directing new updates toward previous low-rank subspaces.
By integrating these components, ReLoRA effectively trains large-scale transformer language models with up to 1.3 billion parameters, achieving performance comparable to traditional full-rank training methods. Notably, ReLoRA offers significant resource savings, reducing memory usage by up to 5.5 GB per GPU and enhancing training speed by 9–4\%, depending on model size and hardware configuration. These advantages make ReLoRA a promising approach for efficient large-scale model training.

\section{Methodology}
\label{sec:methodology}

\subsection{Data Preparation}
\label{subsec:data-preparation}
To process large markdown documents, like the study materials used for training in the IC exam, we employ a technique called Dynamic Markdown Chunking. This technique involves splitting markdown files into smaller, contextually coherent sections by leveraging the structure provided by the document’s headers. While keeping the chunk size within predefined token limits, we think that this method would help us retains relevant information grouped under a common topic.

The Dynamic Markdown Chunking method follows a two-step approach to ensure each chunk is meaningful and remains within the model's token limit:
\begin{enumerate}
    \item \textbf{Initial Chunking by Header Level}:
    \begin{itemize}
        \item The document is initially chunked based on the hierarchy of headers, starting with level 2 headers (\#\#). Each \#\# header begins a new chunk, and all subsequent content is included until the next \#\# header. Higher-level headers (\#) are retained to provide context, while deeper-level headers (\#\#\#, \#\#\#\#) are included within the chunk.
        \item For example, a chunk beginning with a \#\# header would include all preceding \# headers and subsequent content under deeper headers (e.g., \#\#\#) until the next \#\# header.
    \end{itemize}
    \item \textbf{Further Splitting of Large Chunks}:
    \begin{itemize}
        \item If a chunk exceeds the token limit, it is further split based on deeper header levels (e.g., \#\#\#, \#\#\#\#) or logical content divisions such as paragraphs or bullet points. Large paragraphs and lists are broken down carefully to preserve context.
        \item For example, if the content under a header is too long, it may be split into smaller sub-chunks by breaking paragraphs or bullet points while retaining their logical flow.
    \end{itemize}
\end{enumerate}

This method enables the model to handle long documents like the IC exam study materials by splitting them into manageable chunks. The main goal is to ensure that the model can processes the given chunk of documents, while can also understand the relevance between common chunks.

\subsection{Data Augmentation}
\label{subsec:data-augmentation}

Our training dataset includes a limited number of mock exams and over 1.3 million tokens of study materials. However, this is relatively small compared to the extensive range of topics and scenarios covered in the IC exam. To bridge this gap and ensure the model's performance remains robust, we implemented several data augmentation techniques. These methods enhance how knowledge is connected and applied, thereby improving the model’s ability to generalize effectively. The following subsections detail the various augmentation techniques employed.

\subsubsection{Self-Supervised Data Augmentation using Biased Zero-Shot}
\label{subsubsec:self-supervised-augmentation}

To generate reasoning data for exam questions with non-descriptive answers, we configure the model to produce each answer choice along with a supporting reason explaining why that choice could be correct. The reasoning for the correct answer is utilized during SFT, while the reasons for incorrect answers serve as negative examples in DPO \cite{NEURIPS2023_a85b405e}.

\subsubsection{Multiple System Prompts Augmentation}
\label{subsubsec:multiple-system-prompts}

To diversify the training data, we employed multiple system prompts, each designed to simulate the varied ways users might interact with the language model. By presenting the same IC exam content across different contexts, these prompts expose the model to a broader spectrum of scenarios. This approach enhances the model's ability to generalize and adapt, ensuring it can effectively respond across diverse financial domains.

\subsubsection{Multiple-Choice Shuffling}
\label{subsubsec:mc-shuffling}

To mitigate positional bias in the model's learning, we implemented Multiple-Choice Shuffling. This technique randomizes the order of answer options. The objective is to ensure that the model concentrates on the content of the question rather than learning the choice patterns. This approach aligns with findings from the study ``Leveraging Large Language Models for Multiple Choice Question Answering'' \cite{robinson2023leveraginglargelanguagemodels}, which emphasizes the significance of how answer choices are presented to models and suggests that varying their order can serve as an effective augmentation strategy.

\subsubsection{Multi-LLM Response Generation and Validation for DPO Training}
\label{subsubsec:multi-llm-validation}

To enhance response diversity and accuracy, we utilized multiple LLMs, including our base model, to generate responses for each question using both zero-shot and Chain-of-Thought (CoT) prompts \cite{NEURIPS2022_9d560961}. Each prompt type generated distinct reasoning patterns, which were validated separately. Correct responses were then used in DPO training, and incorrect responses were labeled as rejected.

Using multiple prompt types introduces a variety of responses for each question, enriching the dataset and improving the model’s performance. The prompt types employed include zero-shot prompts, which generate immediate answers followed by post-answer reasoning, and CoT prompts, which guide the model to break down reasoning steps before producing an answer.
After generating multiple responses from various LLMs for each prompt type and question, we pair the responses and validate them. The response that adheres to the correct format and provides the correct answer is labeled as accepted, while the other is labeled as rejected.

\subsubsection{Question-Answer Generation from Markdown}
\label{subsubsec:qa-generation-markdown}

We augmented the dataset by generating question-answer pairs directly from markdown documents. This technique leverages the structured nature of markdown files, particularly headers, to create relevant questions and answers. Headers (\#, \#\#, \#\#\#) represent key sections or concepts, making them ideal for generating questions. The corresponding content under each header serves as the answer, resulting in a large set of meaningful question-answer pairs.

This process enhances the dataset by mimicking real-world question styles and helping the model establish better connections between topics and explanations. By focusing on question-answer pairs derived from the markdown’s structure, the model gains a deeper understanding of hierarchical information and improves its performance in tasks such as IC exam preparation.

\subsection{Model Optimization Techniques}

\subsubsection{Continual Pre-Training}
We applied CPT \cite{ke2023continualpretraininglanguagemodels} to train the model on a subset of the study materials using the Chunked Markdown data. The goal was to help the model gain foundational knowledge in finance by exposing it to structured content related to financial products, regulations, and complex financial instruments.

\subsubsection{Supervised Fine-Tuning}
We employed two different approaches for SFT to improve reasoning and question-answering capabilities:

\begin{itemize}
    \item \textbf{CoT on Bias-Generated Reasoning}: This fine-tuning helped the model improve its reasoning abilities by generating explanations for correct answers in a CoT format.
    \item \textbf{Question-Answer Fine-Tuning}: The model was fine-tuned on a variety of question-answer pairs, which included markdown-generated Q\&A, bias-generated CoT and zero-shot reasoning, and multiple-choice shuffling. This enabled the model to generalize across different question structures and domains, resulting in improved performance.
\end{itemize}

\subsubsection{Direct Preference Optimization}
Two variations of DPO \cite{NEURIPS2023_a85b405e} were applied to refine the model’s decision-making and reasoning capabilities:

\begin{itemize}
    \item \textbf{CoT on Bias-Generated Reasoning}: This technique trained the model to generate and prioritize correct explanations for biased answers, discarding incorrect explanations.
    \item \textbf{Zero-shot Multiple-Choice Shuffling and Multi-LLM Response Generation}: Using multiple-choice shuffling, the model was trained to prioritize correct answers based on content rather than the position of the answer. Additionally, using responses from various LLMs, the model was trained to prioritize correct answers with correct answer template over incorrect ones.
\end{itemize}

\subsection{Parameter-Efficient Fine-Tuning Techniques}
\label{subsec:param-eff-ft-techniques}

\subsubsection{Rank-stabilized LoRA}
For CPT \cite{ke2023continualpretraininglanguagemodels}, we utilized RsLoRA \cite{kalajdzievski2023rankstabilizationscalingfactor}, which stabilized rank during low-rank adaptations, ensuring efficient fine-tuning of the model’s foundational financial knowledge.

\subsubsection{ReLoRA}
We applied ReLoRA \cite{lialin2023relorahighranktraininglowrank}, an iterative approach to training low-rank LoRA modules. Our training objective for each LoRA includes:
\begin{itemize}
    \item CPT on Chunked Markdown for foundational knowledge.
    \item CPT focusing on P2 materials for better handling of complex products.
    \item SFT with CoT Bias-Generated Reasoning.
    \item SFT on markdown-generated Q\&A, CoT and Zero-shot Bias-Generated Reasoning, and multiple-choice shuffling.
    \item DPO on CoT Bias-Generated Reasoning.
    \item DPO on Multiple-Choice Shuffling and Multi-LLM Response Generation.
\end{itemize}

\section{Experimental Setup}

We conducted a preliminary evaluation of foundational instruction-tuned generalist models on the public IC exams.
This evaluation enabled us to assess the model's performance across various sections of the IC exam and to design targeted fine-tuning routines aimed at addressing areas that require further improvement.

We benchmarked various models, including commercially available APIs (i.e., gemini-1.5-flash-002, gemini-1.5-pro-002 \cite{geminiteam2024gemini15unlockingmultimodal}, gpt-4-turbo-2024-04-09, and gpt-4o-2024-11-20 \cite{openai2024gpt4technicalreport}) and instruction-tuned foundational LLMs (i.e., Llama3.1-BB-Instruct \cite{dubey2024llama3herdmodels}, Qwen2-7B-Instruct \cite{yang2024qwen2technicalreport}, SeaLLMs-v3-7B-Chat \cite{damonlp2024seallm3}, and our previously released model, THaLLE-0.1-7B-fa \cite{labs2024thalletexthyperlocallyaugmented}), against the public IC practice exams. Our in-house model is fine-tuned based on the Qwen2-7B-Instruct \cite{yang2024qwen2technicalreport} open-source model, selected for its general performance and instruction-following capabilities.
The training involved specialized data augmentation and optimization techniques to adapt the model to domain-specific tasks in financial consulting, as detailed in Sections~\ref{subsec:data-augmentation}~and~\ref{subsec:training-data}.

\subsection{Dataset}
\label{sec:dataset}
\subsubsection{Training Dataset}
\label{subsec:training-data}

Our training dataset consists of two primary components: mock exams and study materials.

\begin{itemize}
    \item \textbf{Mock Exams}: The dataset includes a limited number of mock exams, each composed of four-choice multiple-choice questions. These exams closely simulate the actual IC exam format and cover content required for all three levels.
    \item \textbf{Study Materials}: In addition to the mock exams, the dataset includes 1,365,594 tokens of study materials in markdown format. These materials encompass a wide range of topics pertinent to each level of the IC exam. The number of tokens of each level of IC exam is shown in Table~\ref{tab:icmddata}. The markdown format ensures the model can parse and understand key sections like definitions, bullet points, and regulations, improving knowledge retrieval and context-aware responses.
\end{itemize}

\begin{table}[h]
\centering
\begin{tblr}{
  colspec = {l r},
  hline{1,2,5,6} = {-}{},
}
\textbf{Exam}          & \textbf{Number of Tokens} \\
P1     & 884,542                   \\
P2 & 153,650                   \\
P3 & 327,402                   \\
\textbf{Total}         & \textbf{1,365,594}
\end{tblr}
    \caption{Number of tokens in study materials datasets, separated by levels.}
    \label{tab:icmddata}
\end{table}

\subsection{Public Investment Consultant Practice Exam}
\label{subsec:ic-exam}

For the test data, we utilized publicly available practice exams provided directly by SET. These exams are specifically designed to prepare candidates for the IC exams and are accessible on SET official website. The practice exams selected for benchmarking include:

\begin{itemize}
    \item P1: Basic investment products such as equities and bonds \cite{SET2022PracticePlain}.
    \item P2: Advanced instruments such as structured bonds and complex mutual funds \cite{SET2022PracticeComplexP2}.
    \item P3: Focuses on derivatives like futures, options, and other financial derivatives \cite{SET2022PracticeComplexP3}.
\end{itemize}

The exams cover three key levels, as shown in Table~\ref{tab:publicicexam}, each focusing on distinct aspects of financial product knowledge. We utilized these exams as the test dataset to facilitate direct comparisons across models and ensure easy reproducibility of the benchmarking process. The evaluation was conducted to assess the model’s understanding and performance on each task.

\begin{table}[h]
\centering
\begin{tblr}{
  colspec = {l r r},
  hline{1,2,5} = {-}{},
}
\textbf{Exam}          & \textbf{\# Problems} & \textbf{Passing Criteria} \\
P1     & 50                  & 70\%                      \\
P2 & 25                  & 70\%                      \\
P3 & 25                  & 70\%                      \\
\end{tblr}
\caption{Number of problems in public IC practice exams, separated by levels.}
\label{tab:publicicexam}
\end{table}

\section{Experimental Results}
\label{sec:results}

The evaluation results, presented in Table~\ref{tab:results}, reveal a competitive performance landscape among the evaluated models on the public IC practice exams (P1, P2, and P3). Among the commercial APIs, gpt-4o-2024-11-20 \cite{openai2024gpt4technicalreport} consistently achieves the highest scores across all three tests, tying with gemini-1.5-pro-002 \cite{geminiteam2024gemini15unlockingmultimodal} in P1 and P2, while matching gpt-4-turbo-2024-04-09 in P3. This suggests that these models, particularly gpt-4o-2024-11-20 \cite{openai2024gpt4technicalreport}, exhibit strong generalization and adaptability across diverse exam scenarios.

In the category of open instruction-tuned foundational LLMs, performance is notably more varied. Models such as Llama3.1-8B-Instruct \cite{dubey2024llama3herdmodels} and THaLLE-0.1-7B-fa \cite{labs2024thalletexthyperlocallyaugmented} demonstrate moderate proficiency, with THaLLE-0.1-7B-fa outperforming others in this category. However, the standout performance among open models comes from THaLLE-0.1-7B-ic, which achieves parity with top-performing commercial models, scoring 72\% on P1 and P2 and 84\% on P3. This underscores the efficacy of domain-specific fine-tuning (as indicated by the ``IC'' variant) in narrowing the performance gap between open-source and commercial models.

A notable trend in the results is the performance disparity between commercial APIs and open foundational models. Commercial models generally display greater robustness and consistency, reflecting their extensive training on diverse and high-quality data. However, the success of THaLLE-0.1-7B-ic illustrates the potential of targeted instruction-tuning strategies in enabling open models to rival their commercial counterparts.

Another observation is the variability in results across different exam sections (P1, P2, and P3). Models such as gpt-4-turbo-2024-04-09 and THaLLE-0.1-7B-ic excel in P3, a section that may demand higher reasoning or specialized knowledge, whereas other models show relative declines in performance. This suggests that some models may be better optimized for specific task types, highlighting the importance of aligning model training objectives with the domain-specific requirements of the evaluation.

In conclusion, the results emphasize the continued dominance of commercial APIs in general-purpose tasks but also showcase the promise of fine-tuned, open-source solutions like THaLLE-0.1-7B-ic in achieving competitive performance within specific domains. This finding encourages further exploration of instruction-tuning and targeted fine-tuning approaches to bridge the performance gap in cost-effective and open-access settings.

\begin{table}[h]
    \centering
    \begin{tabular}{l >{\centering\arraybackslash}p{2.5cm} >{\centering\arraybackslash}p{2.5cm} >{\centering\arraybackslash}p{2.5cm}}
        \hline
        \multirow{2}{*}{\textbf{Model}}                                       & \multicolumn{3}{c}{\textbf{Public IC Practice Exams}} \\
                                                                              & \textbf{P1}   & \textbf{P2}   & \textbf{P3}   \\
        \hline
        \multicolumn{4}{l}{\textbf{Commercial APIs}}                                    \\
        gemini-1.5-flash-002 \cite{geminiteam2024gemini15unlockingmultimodal} & 68\%          & 56\%          & 72\%          \\
        gemini-1.5-pro-002 \cite{geminiteam2024gemini15unlockingmultimodal}   & \textbf{74\%} & \textbf{76\%} & 72\%          \\
        gpt-4-turbo-2024-04-09                                                & 72\%          & 68\%          & \textbf{84\%} \\
        gpt-4o-2024-11-20 \cite{openai2024gpt4technicalreport}                              & \textbf{74\%} & \textbf{76\%} & \textbf{84\%} \\
        \hline
        \multicolumn{4}{l}{\textbf{Open Instruction-Tuned Foundational LLMs}}           \\
        Llama3.1-8B-Instruct \cite{dubey2024llama3herdmodels}                 & 54\%          & 64\%          & 48\%          \\
        Qwen2-7B-Instruct \cite{yang2024qwen2technicalreport}                 & 64\%          & 48\%          & 56\%          \\
        SeaLLMs-v3-7B-Chat \cite{damonlp2024seallm3}                          & 50\%          & 28\%          & 48\%          \\
        THaLLE-0.1-7B-fa \cite{labs2024thalletexthyperlocallyaugmented}       & 64\%          & 52\%          & 60\%          \\
        \hline
        \textbf{THaLLE-0.1-7B-ic}                                             & \textbf{72\%} & \textbf{72\%} & \textbf{84\%} \\
        \hline
    \end{tabular}
    \caption{Model performance on Public Investment Consulting Practice Exams (P1, P2, and P3).}
    \label{tab:results}
\end{table}

\section{Conclusion}
\label{sec:conclusion}

In this work, we present THaLLE-IC, an extended version of the IC variants in the THaLLE project, specifically fine-tuned for the IC exam. The model leverages a combination of CPT, SFT, and DPO to adapt effectively to the demands of financial advisory tasks. To address the limited size of the training dataset, advanced data augmentation techniques, such as multiple-choice shuffling and multi-LLM response generation, were employed. These methods played a crucial role in enhancing the model's performance, enabling it to generalize across diverse financial scenarios.

THaLLE-IC exceeded the 70\% passing threshold across all IC exams (P1, P2, and P3).
Its performance on P3, matching GPT-4 models \cite{openai2024gpt4technicalreport} at 84\%, highlights the model’s potential to approach the capabilities of state-of-the-art commercial APIs in complex financial reasoning and comprehension tasks.
These results validate the effectiveness of domain-specific fine-tuning and data augmentation strategies in building models for specialized applications.

While commercial models like GPT-4-turbo-2024-04-09 exhibited greater consistency across all exams, the success of THaLLE-IC emphasizes the growing potential of open-source solutions in achieving comparable performance at a reduced cost. This work demonstrates that by tailoring training methodologies to the specific needs of financial certification and advisory tasks, open-source models can be highly effective.

\section{Contributions and Acknowledgments}

We extend our gratitude to the executive team for their leadership and support.
\newline
\newline
\textit{Core Contributors:} Atthakorn Petchsod, Pornchanan Balee, Danupat Khamnuansin, Anuruth Lertpiya, Chanatip Saetia
\newline
\textit{Project Managers:} Apinya Thianthong
\newline
\textit{Executive Leadership\footnote{\label{footnote:name}sort by alphabetical order}:} Monchai Lertsutthiwong, Tawunrat Chalothorn, Thadpong Pongthawornkamol

\newpage
\bibliographystyle{unsrt}
\bibliography{ref}

\newpage

\section*{Appendix}
\appendix
\renewcommand*{\thesection}{\Alph{section}}

\section{Dynamic Markdown Chunking Example}
\label{sec:appendix-md-chunking}

Consider a markdown document with the following structure:
\begin{samepage}
    \begin{verbatim}
# Example Heading 1

## Example Heading 1.1
Text under example heading 1.1.

### Example Heading 1.1.1
Details under example heading 1.1.1.

### Example Heading 1.1.2
Details under example heading 1.1.2.

## Example Heading 1.2
Text under example heading 1.2.

### Example Heading 1.2.1
Details under example heading 1.2.1.

### Example Heading 1.2.2
Details under example heading 1.2.2.
    \end{verbatim}
\end{samepage}

\textbf{Step 1: Initial Chunking by \#\# Headers:}
\begin{itemize}
    \item \textbf{Chunk 1}:
    \begin{samepage}
    \begin{verbatim}
# Example Heading 1

## Example Heading 1.1
Text under example heading 1.1.

### Example Heading 1.1.1
Details under example heading 1.1.1.

### Example Heading 1.1.2
Details under example heading 1.1.2.
    \end{verbatim}
    \end{samepage}
    \item \textbf{Chunk 2}:
    \begin{samepage}
    \begin{verbatim}
# Example Heading 1

## Example Heading 1.2
Text under example heading 1.2.

### Example Heading 1.2.1
Details under example heading 1.2.1.

### Example Heading 1.2.2
Details under example heading 1.2.2.
    \end{verbatim}
    \end{samepage}
\end{itemize}

Each chunk includes all subsections under its main \#\# header, along with the top-level \# header for context.

\textbf{Step 2: Further Splitting if Chunk Exceeds Token Limit:}
\begin{itemize}
    \item If a chunk, such as ``Example Heading 1.2,'' exceeds the token limit, it is further divided into smaller chunks. For example:
    \begin{itemize}
        \item \textbf{Sub-chunk 1}:
        \begin{verbatim}
# Example Heading 1

## Example Heading 1.2
Text under example heading 1.2.
        \end{verbatim}
        \item \textbf{Sub-chunk 2}:
        \begin{verbatim}
# Example Heading 1

## Example Heading 1.2

### Example Heading 1.2.1
Details under example heading 1.2.1.
        \end{verbatim}
        \item \textbf{Sub-chunk 3}:
        \begin{verbatim}
# Example Heading 1

## Example Heading 1.2

### Example Heading 1.2.2
Details under example heading 1.2.2.
        \end{verbatim}
    \end{itemize}
\end{itemize}

\section{System Prompt}
\label{apx:prompt}

\begin{table}[H]
    \centering
    \begin{tabular}{|p{3.25cm}|p{11.5cm}|}
        \hline
        Type                                                   & System Prompt \\
        \hline
        CoT                                       & You are a Certified Thai Investment Consultant (IC) taking a multiple choice exam.\newline Think step-by-step and then finish your answer with "\foreignlanguage{thai}{ดังนั้น คำตอบที่ถูกต้องคือ}: " followed by the correct choice name (1, 2, 3, or 4). \\
        \hline
        Zero-Shot \newline (for bias generation)               & You are a Certified Thai Investment Consultant (IC) taking a test to evaluate your knowledge.\newline Multiple choices question along with four possible answers (1, 2, 3, and 4) will be given to you.\newline Your task is to indicate the correct answer and provide the backup reason.\newline Begin your answer with "\foreignlanguage{thai}{คำตอบที่ถูกต้องคือ}: " followed by the correct choice and then finish your answer with "\textbackslash\textbackslash n\textbackslash\textbackslash n\foreignlanguage{thai}{เหตุผล}:\textbackslash\textbackslash n". \\
        \hline
        Zero-Shot \newline (for Multi-LLM response generation) & You are a Certified Thai Investment Consultant (IC) taking a test to evaluate your knowledge.\newline Multiple choices question along with four possible answers (1, 2, 3, and 4) will be given to you.\newline Your task is to indicate the correct answer and provide the backup reason. \\
        \hline
    \end{tabular}
    \caption{Prompts used in our experiments.}
    \label{tab:system-prompt}
\end{table}

\section{Model Output Examples}
\label{apx:outputs}

Table~\ref{tab:outputs} shows examples model output when given ``CoT'' prompt on a Practice IC exam question.

System Prompt: \\
You are a Certified Thai Investment Consultant (IC) taking a multiple choice exam. \\
Think step-by-step and then finish your answer with "\foreignlanguage{thai}{ดังนั้น คำตอบที่ถูกต้องคือ}: " followed by the correct choice name (1, 2, 3, or 4).

\textnormal{User Prompt:} \\
Question: \foreignlanguage{thai}{ผู้ลงทุนจัดสรรเงินลงทุนหลักทรัพย์ A 40\% ส่วนที่เหลือจัดสรรเงินลงทุนในหลักทรัพย์ B หากอัตราผลตอบแทนที่คาดหวังของหลักทรัพย์ A และหลักทรัพย์ B เท่ากับ 30\% และ 24\% ตามลำดับ อัตราผลตอบแทนที่คาดหวังของกลุ่มหลักทรัพย์นี้จะเท่ากับเท่าใด \\
\\
1) 24.0\% \\
2) 26.4\% \\
3) 27.6\% \\
4) 30.0\%}

\pagebreak

\begin{longtable}{|p{3.7cm}|L{11.3cm}|}
        \hline
        \textbf{Model}                         & \textbf{Output} \\
        \hline
        gemini-1.5-flash-002       & \foreignlanguage{thai}{ขั้นตอนที่ 1: คำนวณสัดส่วนการลงทุนในหลักทรัพย์ B}\newline

\foreignlanguage{thai}{เนื่องจากหลักทรัพย์ A มีสัดส่วนการลงทุน 40\% ดังนั้นสัดส่วนการลงทุนในหลักทรัพย์ B จะเท่ากับ 100\% - 40\% = 60\%}\newline

\foreignlanguage{thai}{ขั้นตอนที่ 2: คำนวณอัตราผลตอบแทนที่คาดหวังของกลุ่มหลักทรัพย์}\newline

\foreignlanguage{thai}{อัตราผลตอบแทนที่คาดหวังของกลุ่มหลักทรัพย์ = (สัดส่วนการลงทุนในหลักทรัพย์ A × อัตราผลตอบแทนที่คาดหวังของหลักทรัพย์ A) + (สัดส่วนการลงทุนในหลักทรัพย์ B × อัตราผลตอบแทนที่คาดหวังของหลักทรัพย์ B)}\newline

\foreignlanguage{thai}{อัตราผลตอบแทนที่คาดหวังของกลุ่มหลักทรัพย์ = (40\% × 30\%) + (60\% × 24\%) = 12\% + 14.4\% = 26.4\%}\newline

\foreignlanguage{thai}{ดังนั้น คำตอบที่ถูกต้องคือ: 2} \\
        \hline
        gemini-1.5-pro-002         & \foreignlanguage{thai}{ขั้นตอนที่ 1: คำนวณสัดส่วนการลงทุนในหลักทรัพย์ B}\newline

\foreignlanguage{thai}{หลักทรัพย์ A มีสัดส่วนการลงทุน 40\% ดังนั้น หลักทรัพย์ B จะมีสัดส่วนการลงทุน 100\% - 40\% = 60\%}\newline

\foreignlanguage{thai}{ขั้นตอนที่ 2: คำนวณอัตราผลตอบแทนที่คาดหวังของกลุ่มหลักทรัพย์}\newline

\foreignlanguage{thai}{อัตราผลตอบแทนที่คาดหวังของกลุ่มหลักทรัพย์ = (สัดส่วนการลงทุนในหลักทรัพย์ A × อัตราผลตอบแทนที่คาดหวังของหลักทรัพย์ A) + (สัดส่วนการลงทุนในหลักทรัพย์ B × อัตราผลตอบแทนที่คาดหวังของหลักทรัพย์ B)}\newline

\[
= (40\% \times 30\%) + (60\% \times 24\%)
\]
\[
= 12\% + 14.4\%
\]
\[
= 26.4\%
\]

\foreignlanguage{thai}{ดังนั้น คำตอบที่ถูกต้องคือ: 2} \\
        \hline
        gpt-4-turbo-2024-04-09     & \foreignlanguage{thai}{จากข้อสอบ ผู้ลงทุนจัดสรรเงินลงทุนในหลักทรัพย์ A 40\% และในหลักทรัพย์ B 60\% (เพราะเหลือจาก 100\% - 40\% = 60\%)}\newline

\foreignlanguage{thai}{โดยหลักทรัพย์ A มีอัตราผลตอบแทนที่คาดหวังเท่ากับ 30\%}\newline
\foreignlanguage{thai}{และหลักทรัพย์ B มีอัตราผลตอบแทนที่คาดหวังเท่ากับ 24\%}\newline

\foreignlanguage{thai}{เพื่อหาอัตราผลตอบแทนรวมของกลุ่มหลักทรัพย์ที่คาดหวัง สามารถคำนวณได้จาก:}\newline
\foreignlanguage{thai}{(อัตราจัดสรรหลักทรัพย์ A × อัตราผลตอบแทนหลักทรัพย์ A) + (อัตราจัดสรรหลักทรัพย์ B × อัตราผลตอบแทนหลักทรัพย์ B)}\newline

\foreignlanguage{thai}{นั่นคือ:}\newline
\[
(0.40 \times 30\%) + (0.60 \times 24\%)
\]
\[
= 0.12 + 0.144
\]
\[
= 0.264 \foreignlanguage{thai}{หรือ} 26.4\%
\]

\foreignlanguage{thai}{ดังนั้น คำตอบที่ถูกต้องคือ: 2} \\
        \hline
        gpt-4o-2024-11-20          & \#\#\# \foreignlanguage{thai}{การคำนวณ:}\newline
\foreignlanguage{thai}{การคำนวณอัตราผลตอบแทนที่คาดหวังของพอร์ตการลงทุน} **(Expected Return of Portfolio)** \foreignlanguage{thai}{สามารถคำนวณได้จากสูตรดังนี้:}\newline

\[
E(R_p) = w_A \times E(R_A) + w_B \times E(R_B)
\]

\foreignlanguage{thai}{โดย:}\newline
- \( w_A \) \foreignlanguage{thai}{และ} \( w_B \) \foreignlanguage{thai}{คือสัดส่วนเงินลงทุนในหลักทรัพย์ A และ B (เป็นทศนิยม)}\newline
- \( E(R_A) \) \foreignlanguage{thai}{และ} \( E(R_B) \) \foreignlanguage{thai}{คืออัตราผลตอบแทนที่คาดหวังของหลักทรัพย์ A และ B}\newline

\#\#\# \foreignlanguage{thai}{ขั้นตอนการคำนวณ:}\newline
1. \textbf{\foreignlanguage{thai}{สัดส่วนการลงทุน}}:\newline
   - \foreignlanguage{thai}{สัดส่วนในหลักทรัพย์ A} (\( w_A \)) \foreignlanguage{thai}{= 40\% = 0.40}\newline
   - \foreignlanguage{thai}{สัดส่วนในหลักทรัพย์ B} (\( w_B \)) \foreignlanguage{thai}{= 100\% - 40\% = 60\% = 0.60}\newline

2. \textbf{\foreignlanguage{thai}{อัตราผลตอบแทนที่คาดหวัง}}:\newline
   - \foreignlanguage{thai}{ผลตอบแทนที่คาดหวังของหลักทรัพย์ A} (\( E(R_A) \)) \foreignlanguage{thai}{= 30\% = 0.30}\newline
   - \foreignlanguage{thai}{ผลตอบแทนที่คาดหวังของหลักทรัพย์ B} (\( E(R_B) \)) \foreignlanguage{thai}{= 24\% = 0.24}\newline

3. \textbf{\foreignlanguage{thai}{แทนค่าในสูตร}}:\newline
   \[
   E(R_p) = w_A \times E(R_A) + w_B \times E(R_B)
   \]
   \[
   = (0.40 \times 0.30) + (0.60 \times 0.24)
   \]
   \[
   = 0.12 + 0.144
   \]
   \[
   = 0.264 = 26.4\%
   \]

\#\#\# \foreignlanguage{thai}{สรุป:}\newline
\foreignlanguage{thai}{ดังนั้น คำตอบที่ถูกต้องคือ:} **2** \\
        \hline
        Meta-Llama-3.1-8B-Instruct & \foreignlanguage{thai}{ในการคำนวณอัตราผลตอบแทนที่คาดหวังของกลุ่มหลักทรัพย์ เราต้องคำนวณผลตอบแทนที่คาดหวังของแต่ละหลักทรัพย์แล้วรวมกัน} \newline\newline \foreignlanguage{thai}{ผลตอบแทนที่คาดหวังของหลักทรัพย์} A \foreignlanguage{thai}{คือ} 40\% \foreignlanguage{thai}{ของเงินลงทุน และอัตราผลตอบแทนที่คาดหวังของหลักทรัพย์} A \foreignlanguage{thai}{คือ} 30\% \foreignlanguage{thai}{ดังนั้น ผลตอบแทนที่คาดหวังของหลักทรัพย์} A \foreignlanguage{thai}{คือ} 0.4 x 30\% = 12\% \newline\newline \foreignlanguage{thai}{ผลตอบแทนที่คาดหวังของหลักทรัพย์} B \foreignlanguage{thai}{คือ} 60\% \foreignlanguage{thai}{ของเงินลงทุน} (100\% - 40\%) \foreignlanguage{thai}{และอัตราผลตอบแทนที่คาดหวังของหลักทรัพย์} B \foreignlanguage{thai}{คือ} 24\% \foreignlanguage{thai}{ดังนั้น ผลตอบแทนที่คาดหวังของหลักทรัพย์} B \foreignlanguage{thai}{คือ} 0.6 x 24\% = 14.4\% \newline\newline \foreignlanguage{thai}{อัตราผลตอบแทนที่คาดหวังของกลุ่มหลักทรัพย์ คือผลรวมของผลตอบแทนที่คาดหวังของหลักทรัพย์} A \foreignlanguage{thai}{และหลักทรัพย์} B \foreignlanguage{thai}{ซึ่งก็คือ} 12\% + 14.4\% = 26.4\% \newline\newline \foreignlanguage{thai}{ดังนั้น คำตอบที่ถูกต้องคือ}: 2) \\
        \hline
        Qwen2-7B-instruct          & \foreignlanguage{thai}{การคำนวณอัตราผลตอบแทนที่คาดหวังของกลุ่มหลักทรัพย์นี้จะใช้สูตรดังนี้}: \newline\newline (40\% x 30\%) + (60\% x 24\%) = \foreignlanguage{thai}{อัตราผลตอบแทนที่คาดหวังของกลุ่มหลักทรัพย์} \newline\newline = 12\% + 14.4\% = 26.4\% \newline\newline \foreignlanguage{thai}{ดังนั้น คำตอบที่ถูกต้องคือ}: 2) 26.4\% \\
        \hline
        SeaLLMs-v3-7B-Chat         & Solution: \foreignlanguage{thai}{ให้สมมติว่ามีเงินลงทุนรวมเป็น} 100 \foreignlanguage{thai}{บาท ดังนั้นเงินลงทุนในหลักทรัพย์} A \foreignlanguage{thai}{จะเป็น} 40 \foreignlanguage{thai}{บาท และเงินลงทุนในหลักทรัพย์} B \foreignlanguage{thai}{จะเป็น} 60 \foreignlanguage{thai}{บาท อัตราผลตอบแทนที่คาดหวังของกลุ่มหลักทรัพย์นี้สามารถคำนวณได้จากสูตร}: (40\% * 30\%) + (60\% * 24\%) = 12\% + 14.4\% = 26.4\% \foreignlanguage{thai}{ดังนั้น คำตอบที่ถูกต้องคือ}: 2) 26.4\% \\
        \hline
        THaLLE-0.1-7B-fa           & \foreignlanguage{thai}{ดังนั้น คำตอบที่ถูกต้องคือ}: 2) 26.4\% \\
        \hline
        THaLLE-0.1-7B-ic           & \foreignlanguage{thai}{การคำนวณอัตราผลตอบแทนที่คาดหวังของกลุ่มหลักทรัพย์นี้สามารถทำได้โดยการนำอัตราผลตอบแทนที่คาดหวังของแต่ละหลักทรัพย์มาคูณกับสัดส่วนของเงิน\hspace{0pt}ลงทุนในแต่ละหลักทรัพย์แล้วนำผลลัพธ์รวมกัน ดังนี้} \newline\newline (0.4 x 30\%) + (0.6 x 24\%) = 12\% + 14.4\% = 26.4\% \newline\newline \foreignlanguage{thai}{ดังนั้น คำตอบที่ถูกต้องคือ}: 2) 26.4\% \\
        \hline
    \caption{Model output examples for IC exam questions.}
    \label{tab:outputs}
\end{longtable}

\end{document}